\documentclass[final]{cvpr}

\usepackage{bm}
\usepackage{multirow}
\usepackage{times}
\usepackage{epsfig}
\usepackage{graphicx}
\usepackage{amsmath}
\usepackage{amssymb}
\usepackage{tabularx}
\usepackage{booktabs}
\usepackage[super]{nth}
\usepackage{setspace}

\newcommand{\nothing}[1]{}
\newcommand{\PV}{PLIVox\xspace}
\newcommand{\PVs}{PLIVoxs\xspace}

\newcommand{\encoder}{\phi_{\mathrm{E}}}    
\newcommand{\decoder}{\phi_{\mathrm{D}}}    
\newcommand{\decmu}{\mu_{\mathrm{D}}}       
\newcommand{\decstd}{\sigma_{\mathrm{D}}}   
\newcommand{\x}{\bm{x}}
\newcommand{\y}{\bm{y}}
\newcommand{\normal}{\bm{n}}
\newcommand{\twist}{\bm{\xi}}

\newcommand{\parahead}[1]{\vspace{0.5mm}\noindent\textbf{#1.}\ }

\newif\ifarxiv

\usepackage[pagebackref=true,breaklinks=true,colorlinks,bookmarks=false]{hyperref}
\usepackage{cleveref}

\crefname{equation}{\text{Eq}}{\text{Eq}}
\crefname{tab}{\text{Tab.}}{\text{Tab.}}
\crefname{fig}{\text{Fig.}}{\text{Fig.}}
\crefname{table}{\text{Tab.}}{\text{Tab.}}
\crefname{figure}{\text{Fig.}}{\text{Fig.}}
\crefname{section}{\text{Sec.}}{\text{Sec.}}

\def\cvprPaperID{1972} 
\def\confYear{CVPR 2021}

{\global\arxivtrue}
\newcommand{\videourl}{\url{https://youtu.be/yxkIQFXQ6rw}}

\begin{document}

\title{DI-Fusion: Online Implicit 3D Reconstruction with Deep Priors}

\author{Jiahui Huang \quad Shi-Sheng Huang \quad Haoxuan Song \quad Shi-Min Hu\thanks{corresponding author.}\\
BNRist, Department of Computer Science and Technology, Tsinghua University,
Beijing
}

\maketitle

\begin{abstract}
Previous online 3D dense reconstruction methods struggle to achieve the balance between memory storage and surface quality, largely due to the usage of stagnant underlying geometry representation, such as TSDF (truncated signed distance functions) or surfels, without any knowledge of the scene priors. 
In this paper, we present DI-Fusion (Deep Implicit Fusion), based on a novel 3D representation, \ie Probabilistic Local Implicit Voxels (\PVs), for online 3D reconstruction with a commodity RGB-D camera. 
Our \PV encodes scene priors considering both the local geometry and uncertainty parameterized by a deep neural network. 
With such deep priors, we are able to perform online implicit 3D reconstruction achieving state-of-the-art camera trajectory estimation accuracy and mapping quality, while achieving better storage efficiency compared with previous online 3D reconstruction approaches. Our implementation is available at \url{https://www.github.com/huangjh-pub/di-fusion}.
\end{abstract}
\vspace{-1em}

\section{Introduction}

Online 3D dense reconstruction has made great progress in the past ten years~\cite{Cao:2018:RHT:3278329.3182157,DBLP:journals/tog/ChenBI13,dai2017bundlefusion,DBLP:journals/tvcg/KahlerPRSTM15,KinectFusion,DBLP:journals/tog/NiessnerZIS13,DBLP:conf/rss/WhelanLSGD15}, enabling a wide range of applications including augmented reality, robotic navigation and games.
Technically most of the previous depth fusion approaches focus on the globally consistent 3D reconstruction with bundle adjustment~\cite{Cao:2018:RHT:3278329.3182157,dai2017bundlefusion} or loop closure~\cite{DBLP:conf/rss/WhelanLSGD15} techniques. 
However, the underlying representation for the 3D scene itself has seldom changed ever since the success of VoxelHashing~\cite{DBLP:journals/tog/NiessnerZIS13} with Signed Distance Function (SDF) integration on a sparse set of voxels~\cite{DBLP:conf/siggraph/CurlessL96}. 
This leads to the drawback of previous depth fusion systems that often costs a huge amount of memory storage even for moderate-sized 3D scenes. 
Besides, the geometric quality could be unsatisfactory with non-complete regions or objects~\cite{DBLP:conf/cvpr/DaiRBRSN18} due to the uncertainties caused by sensor noise or scan ambiguities such as view occlusions.

\begin{figure}[t]
  \centering
  \includegraphics[width=\linewidth]{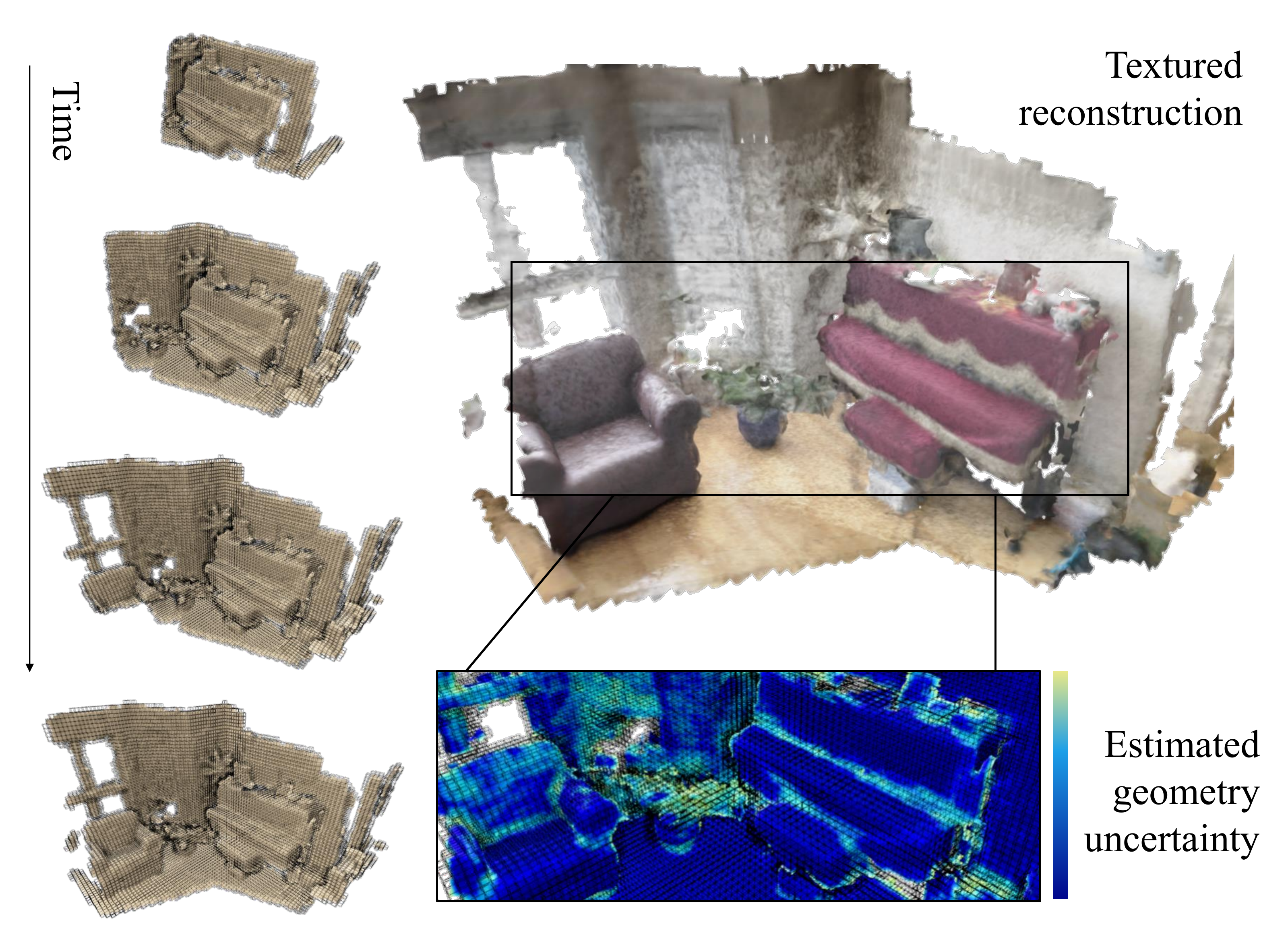}
  \caption{\textbf{DI-Fusion} incrementally builds up a continuous 3D scene from an RGB-D sequence. The tracking and mapping algorithm are fully based on our novel local deep implicit scene representation incorporating learned priors, where both the geometry and its uncertainty are estimated.}
  \label{fig:teaser}
  \vspace{-1.5em}
\end{figure}

On the other hand, recent efforts from the deep geometry learning community have demonstrated the power of implicit geometric representation parameterized by neural networks~\cite{chen2019imnet,mescheder2019occupancy,park2019deepsdf}.
By representing the geometry as a continuous {implicit} function, the underlying shape can be extracted at arbitrary resolution, which introduces more flexibilities.
These methods are also efficient {since} the network structure used to regress implicit function only consists of simple fully connected layers.
Another important feature of such deep implicit representation is the capability to encode geometric priors, which enables many applications such as shape interpolation or reconstruction~\cite{park2019deepsdf}.
This capability can be generalized to scene-level
by decomposing and encoding the implicit fields in local voxels, leading to high-quality reconstruction agnostic to semantics~\cite{chabra2020deepls,jiang2020local}.

The power of such deep implicit representation motivates us to incorporate it into online 3D dense reconstruction systems.
By encoding meaningful scene priors with a continuous function, we can achieve improved surface reconstruction as well as accurate camera trajectory.
However, several challenges need to be overcome before this new representation can be successfully applied in an online fusion scenario:
(1) geometric uncertainty need to be explicitly modeled against sensor noise or view occlusion, 
(2) an accurate camera tracking formulation based on such an implicit representation, which is essential for depth fusion, remains unknown yet, and
(3) an efficient surface mapping strategy that incrementally integrates new observations directly is also missing.

We hence respond with DI-Fusion, the first online 3D reconstruction system with tracking and mapping modules fully supported by deep implicit representations.
To address the above challenges, we first extend the original local implicit grids~\cite{chabra2020deepls,jiang2020local} and adapt it into a novel \textbf{P}robabilistic \textbf{L}ocal \textbf{I}mplicit \textbf{Vox}el (PLIVox), which encodes not only scene geometry but also the \emph{uncertainty} with one deep neural network. We show that such an additional uncertainty encoding is extremely useful during the online depth fusion.
Based on our \PV representation, we devise an approximate gradient for solving the camera tracking problem efficiently.
Moreover, thanks to our tailored encoder-decoder network design, we are able to perform geometry integration on the domain of latent vectors, achieving high quality surface mapping in an efficient way.
We evaluated our approach on public 3D RGB-D benchmark (ICL-NUIM~\cite{handa:etal:ICRA2014}, ScanNet dataset~\cite{dai2017scannet}), showing state-of-the-art or improved tracking and mapping quality compared to previous representations.
We make our implementation publicly available.

\section{Related Works}

\parahead{Online 3D Reconstruction and SLAM}
The success of KinectFusion~\cite{KinectFusion} inspired a lot of works on online 3D reconstruction. 
Early efforts mainly focus on efficient data structures to organize discrete voxels or surfels to enable large-scale 3D reconstruction, such as OctreeFusion~\cite{DBLP:journals/cvgip/ZengZZL13}, VoxelHashing~\cite{DBLP:journals/tog/NiessnerZIS13}, Scalable-VoxelHashing~\cite{DBLP:journals/tog/ChenBI13} and ElasticFusion~\cite{keller2013real,DBLP:conf/rss/WhelanLSGD15}. 
In order to improve the quality and the global consistency of the reconstructed model, subsequent works turn to bundle adjustment~\cite{dai2017bundlefusion,DBLP:journals/tvcg/KahlerPRSTM15,DBLP:journals/ijrr/WhelanKJFLM15} or sub-map multiway registration~\cite{choi2015robust} techniques.
Another line of work couples high-level understanding and geometric modeling including dynamic perception~\cite{newcombe2015dynamicfusion,huang2019clusterslam,huang2020clustervo} or structural regularization~\cite{huang2020wallnet}.
Our work is also relevant to visual SLAM methods~\cite{DBLP:journals/pami/DavisonRMS07,DBLP:conf/eccv/EngelSC14,DBLP:journals/trob/Mur-ArtalT17,DBLP:journals/pami/EngelKC18}, which usually perform camera tracking using sparse features or direct intensity data. 
Readers are referred to~\cite{zollhofer2018state} for a more comprehensive study.

\parahead{Learned Probabilistic Reconstruction}
The advancement of geometry deep learning has recently encouraged the community to incorporate deep neural networks into reconstruction and tracking techniques for more accurate representation of either sparse features (DeepTrack~\cite{Deeptrack2017}), monocular depth observations (CNN-SLAM~\cite{DBLP:conf/cvpr/TatenoTLN17}, CodeSLAM~\cite{DBLP:conf/cvpr/BloeschCCLD18}), or object instances (Fusion++~\cite{DBLP:conf/3dim/McCormacCBDL18}, NodeSLAM~\cite{Sucar:etal:3DV2020}). 
The presence of stochastic sensor noise also motivates many works to consider the probabilistic distribution of the underlying geometry, with either hand-crafted models~\cite{dong2018psdf,yang2020noise} or data-driven ones~\cite{laidlow2019deepfusion,weder2020routedfusion}, where the latter additionally incorporate and explicitly model the approximability of the deep networks~\cite{stutz2020learning}.

\begin{figure}[!t]
  \centering
  \includegraphics[width=\linewidth]{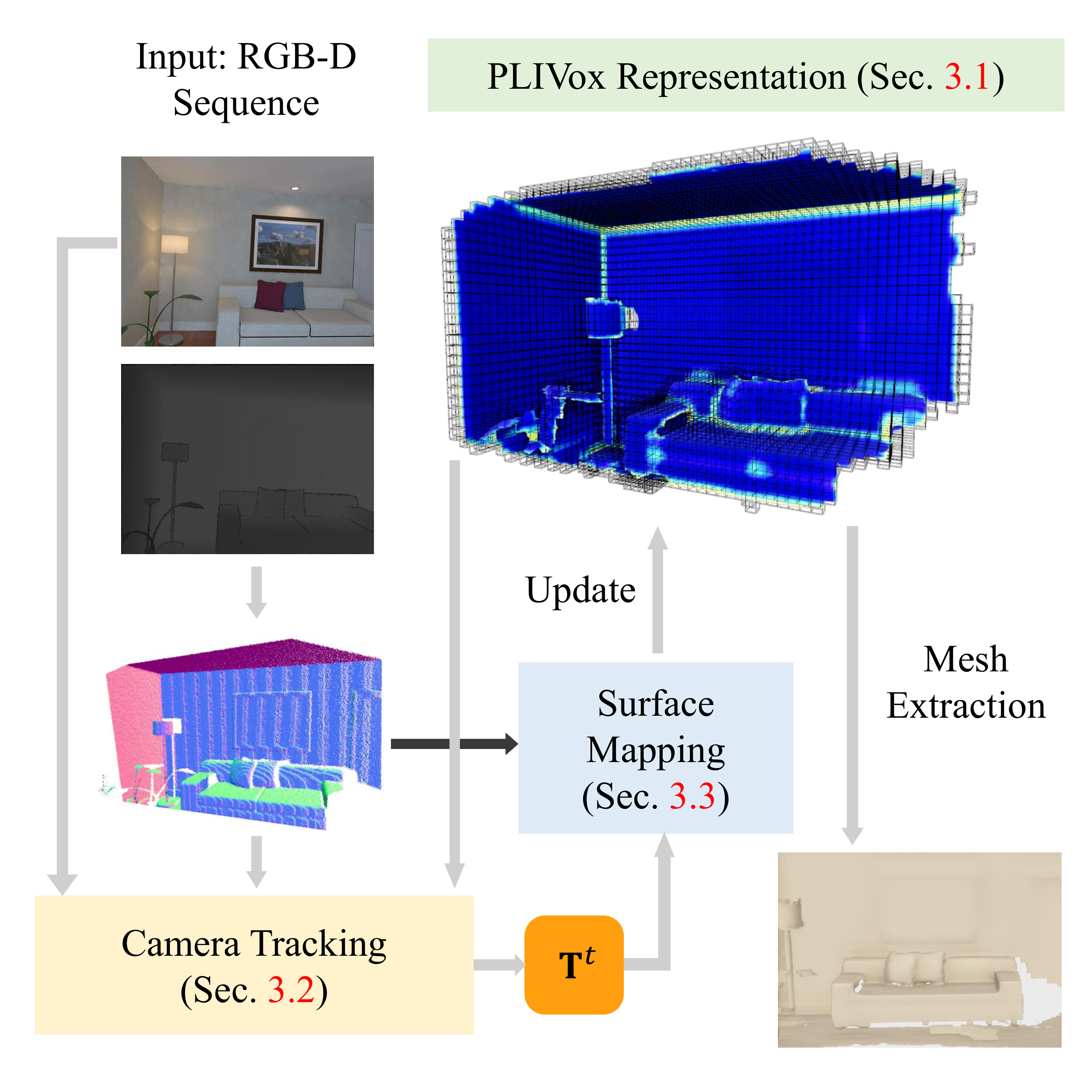}
  \caption{\textbf{Overview}. We represent the reconstructed 3D scene with \PVs~(\cref{subsec:method:plivox}). Given input RGB-D frames, we first estimate the camera pose $\mathbf{T}^t$ by finding the best alignment between the current depth point cloud and the map~(\cref{subsec:method:tracking}), then the depth observations are integrated~(\cref{subsec:method:integration}) for surface mapping. Scene mesh can be extracted any time on demand at any resolution. Note that both the camera tracking and surface mapping are performed directly on the deep implicit representation.}
  \label{fig:overview}
  \vspace{-1em}
\end{figure}

\parahead{Implicit Representation}
The use of implicit function for geometric reconstruction can be dated back to~\cite{DBLP:conf/siggraph/CurlessL96}, where SDF values are stored in a set of occupied voxels describing the surface.
However, even though the implicit function is continuous, the simple discretization~\cite{DBLP:journals/tog/ChenBI13,DBLP:journals/tog/NiessnerZIS13} introduces drawbacks in surface reconstruction quality and memory storage.
As an effort to overcome such drawbacks, \cite{lee2019online,DBLP:journals/ral/MartensPSFS17} propose to model the map using Gaussian Process and perform Bayesian map updates incrementally.
In contrast, the use of implicit representation has triggered much interest in deep learning community these years~\cite{mescheder2019occupancy,park2019deepsdf}. Typical applications include: part segmentation~\cite{chen2019bae}, rendering~\cite{liu2020dist,mildenhall2020nerf}, non-linear fitting~\cite{sitzmann2020implicit}, meta-learning~\cite{sitzmann2020metasdf}, offline reconstruction~\cite{chabra2020deepls,jiang2020local} etc.
Nevertheless, to the best of our knowledge, our DI-Fusion is among the first efforts to incorporate such an implicit representation with deep priors into an online 3D fusion framework.

\section{Method}

\begin{figure}[!t]
  \centering
  \includegraphics[width=\linewidth]{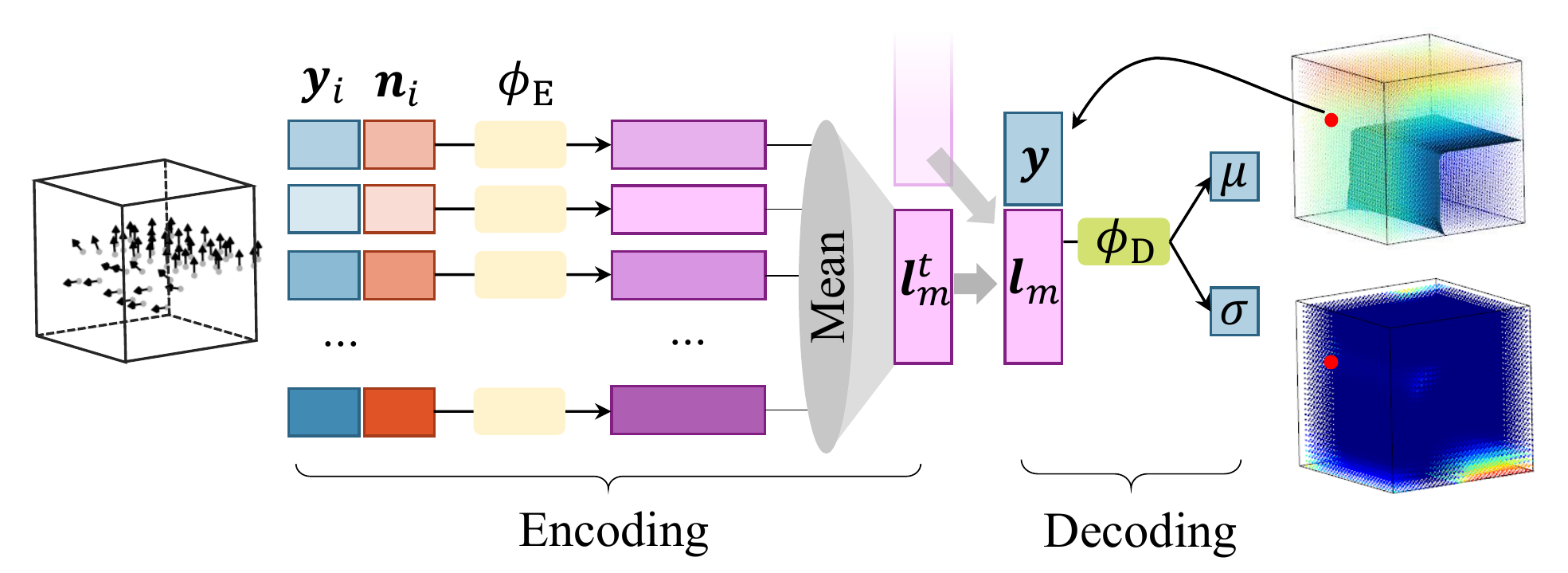}
  \caption{The structure of our encoder-decoder neural network $\Phi$. The encoding and decoding process within one \PV is shown. The arrow between encoding and decoding denotes incremental latent vector update if the latent vector from last frame is available.}
  \label{fig:network}
\end{figure}

\parahead{Overview}
Given a sequential RGB-D stream, our DI-Fusion incrementally builds up a 3D scene based on a novel \textbf{P}robabilistic \textbf{L}ocal \textbf{I}mplicit \textbf{Vox}els (\PV) representation.
Different from previous approaches using discrete voxels without any scene priors, our \PV is implicitly parameterized by a neural network and encodes useful local scene priors effectively (\cref{subsec:method:plivox}).
Based on this representation, we introduce how to robustly perform camera tracking for each frame (\cref{subsec:method:tracking}) and how to perform efficient incremental surface mapping (\cref{subsec:method:integration}).
As a final step mesh can be extracted at arbitrary resolution thanks to the continuous representation, compared to previous approaches which can only reconstruct at a predefined resolution. 
This overview of our DI-Fusion is illustrated in \cref{fig:overview}.

\subsection{\PV Representation}
\label{subsec:method:plivox}

The scene reconstructed is sparsely partitioned into evenly-spaced voxels (\PVs), denoted as $\mathcal{V} = \{ \bm{v}_m = ( \bm{c}_m, \bm{l}_m, w_m ) \}$, with $\bm{c}_m \in \mathbb{R}^3$ being voxel centroid, $ \bm{l}_m \in \mathbb{R}^L $ being the latent vector encoding the scene priors and $w_m \in \mathbb{N}$ being the observation weight.
For an arbitrary point measurement $\x \in R^3$, we can efficiently query its corresponding {\PV} index $m(x)$ using simple division and rounding operations $m(\x): \mathbb{R}^3 \mapsto \mathbb{N}^+$.
The local coordinate of $\x$ in $\bm{v}_{m(\x)}$ is calculated as $\y = \frac{1}{a} (\x-\bm{c}_{m(\x)}) \in [-\frac{1}{2}, \frac{1}{2}]^3$, with $a$ being the voxel size.

\parahead{Probabilistic Signed Distance Function} Different from the previous approaches representing the underlying 3D surface with signed distance function, we represent it using a \emph{probabilistic} signed distance function, where the output at every position $\bm{y}$ is not a SDF but a SDF distribution $s \sim p (\cdot|\bm{y})$. 
In this way, the probabilistic signed distance function encodes the surface geometry and geometric uncertainty at the same time.
Here we model the SDF distribution as a canonical Gaussian distribution $\mathcal{N} (\mu,\sigma^2)$ with $\mu$ and $\sigma$ being the mean and standard deviation respectively. For a more compact representation, we encode the probabilistic signed distance function with a latent vector $\bm{l}_m$ using an encoder-decoder deep neural network $\Phi$.

\parahead{Encoder-Decoder Neural Network} 
The encoder-decoder neural network $\Phi=\{\encoder,\decoder\}$ consists of encoder sub-network $\encoder$ and decoder sub-network $\decoder$ (\cref{fig:network}) that share weights with all \PVs.

The target of the encoder $\encoder$ is to convert the measurements from each depth point observation at frame $t$ to \emph{observation} latent vectors $\bm{l}_m^t$.
Specifically, for all the RGB-D point measurements located in a \PV, $\encoder$ takes in the point measurement's local coordinate $\y$ and normal direction $\normal$, and transforms them to an $L$-dimensional feature vector $\encoder(\y, \normal)$ using only FC~(Fully Connected) layers.
Then the feature vectors from multiple points are aggregated to one latent vector $\bm{l}_m^t$ using a mean-pooling layer (Fig.~\ref{fig:network}).
Here the normal direction $\normal$ is required to eliminate the orientation ambiguity within each {\PV} so the sign of SDF can be inferred by the network. 

For the decoder $\decoder$, the concatenation of the local coordinate {$\y$} and the latent vector $\bm{l}_m$ are taken as input and the output is a 2-tuple $\{\decmu,\decstd\}$, which represents the Gaussian parameters of the probabilistic signed distance function distribution $p (\cdot|\bm{y}) \sim \mathcal{N} (\decmu,\decstd^2)$ at position $\y$. 
Note that the two latent vectors $\bm{l}_m^t$ and $\bm{l}_m$ in $\encoder$ and $\decoder$ are different latent vectors. 
While the \emph{observation} latent vector $\bm{l}_m^t$ encodes the RGB-D observations at frame $t$, the \emph{geometry} latent vector $\bm{l}_m$ fuses the previous $\bm{l}_m^t$ and is stored in each \PV used for decoding. 
Both the observation latent vector and the geometry latent vector have the same dimension, and the geometry latent vector can be updated by the observation latent vector as described in Sec.~\ref{subsec:method:integration}. 

\parahead{Network Training} We train the $\encoder$ and $\decoder$ jointly in an end-to-end fashion, setting $\bm{l}_m^t \equiv \bm{l}_m$.
We adopt the training strategy similar to Conditional Neural Process (CNP~\cite{garnelo2018cnp}) but extend it to the 3D domain.
Specifically, we first build up two training datasets: 
(1) $\mathcal{S} = \{\mathcal{S}_m\}$ for encoder, which is a set of tuples $\mathcal{S}_m=\{(\y_i,\normal_i)\}$ for each PLIVox $\bm{v}_m$ with points $\y_i$ and $\normal_i$ sampled from the scene surface; 
(2) $\mathcal{D} = \{\mathcal{D}_m\}$ for the decoder, which consists of tuples $\mathcal{D}_m =\{(\y_i, s_\mathrm{gt}^i)\}$ where points $\y_i$ are randomly sampled within a PLIVox using a strategy similar to~\cite{park2019deepsdf} with $s_\mathrm{gt}^i$ being the SDF at point $\y_i$. 
We give more details about how to build up such two training datasets in Sec.~\ref{sec:exp}.
During training, we feed $\mathcal{S}_m$ to the encoder for latent vector $\bm{l}_m$ and concatenate the latent vector with each $\y_i$ in $\mathcal{D}_m$ to obtain predicted SDF mean and standard deviation.
The goal of training is to maximize the likelihood of the dataset $\mathcal{D}$ for all training \PVs.
Specifically, the loss function $\mathcal{L}_m$ for each PLIVox $\bm{v}_m$ is written as:
\begin{equation}
\label{equ:loss}
    \mathcal{L}_m = -\sum_{(\y_i, s_\mathrm{gt}^i) \in \mathcal{D}_m} \log \mathcal{N} \left( s_\mathrm{gt}^i; \decmu(\y_i,\bm{l}_m), \decstd^2(\y_i,\bm{l}_m) \right),
\end{equation}
\begin{equation}
    \bm{l}_m = \frac{1}{|\mathcal{S}_m|} \sum_{(\bm{y}_i, \bm{n}_i)\in\mathcal{S}_m} \encoder(\bm{y}_i, \bm{n}_i).
\end{equation}

Additionally, we regularize the norm of the latent vector with a $l_2$-loss which reflects the prior distributions of $\bm{l}_m$.
The final loss function $\mathcal{L}$ we used for training is:
\begin{equation}
\label{equ:final_loss}
    \mathcal{L} = \sum_{\bm{v}_m \in \mathcal{V}} \mathcal{L}_m + \delta \lVert \bm{l}_m \rVert^2.
\end{equation}

\subsection{Camera Tracking}
\label{subsec:method:tracking}

With our \PV encoding of the scene priors including both the scene geometry and uncertainty, we propose a frame-to-model~\cite{KinectFusion} camera tracking method.
We claim that the learned deep priors have enough information of the 3D scene for an accurate camera pose estimation, without the need for extra sparse features as in ~\cite{Cao:2018:RHT:3278329.3182157,dai2017bundlefusion}. 
We propose to formulate the probabilistic signed distance function as an objective function for camera pose estimation, with an approximate gradient for the objective function over camera pose, which makes it converge fast enough during the optimization.
Furthermore, our network is efficient in decoding the probabilistic signed distance function, leading to fast online tracking performance.

\parahead{Tracking} We denote the RGB-D observation at frame $t$ as $\mathcal{O}^t = \{ \mathcal{I}^t , \mathcal{D}^t \}$ with $\mathcal{I}^t$ and $\mathcal{D}^t$ being the intensity and depth data. Given camera intrinsic parameters, the depth measurement $\mathcal{D}^t$ can be re-projected to 3D as point measurements $\mathcal{P}^t = \pi' (\mathcal{D}^t)$, where $\pi$ is the projection function and $\pi'$ is its inverse. Our goal is to estimate $\mathcal{O}^t$'s camera pose $\mathbf{T}^t \in SE(3)$ by optimizing the relative pose $T(\twist^t) = \exp \left( (\twist^t)^\wedge \right)$ ($\twist^t \in se(3)$~\cite{barfoot2017state}) between $\mathcal{O}^t$ and $\mathcal{O}^{t-1}$, \ie $\mathbf{T}^t = \mathbf{T}^{t-1} T(\twist^t)$.
The following objective function is minimized in our system:
\begin{equation}
\label{equ:opt}
\begin{aligned}
E(\twist^t) = E_{\mathrm{sdf}}(\twist^t) + w E_{\mathrm{int}}(\twist^t),
\end{aligned}
\end{equation}
where $E_{\mathrm{sdf}}(\twist^t)$ and $ E_{\mathrm{int}}(\twist^t)$ are the SDF term and intensity term respectively, and $w$ is a weight parameter.
The objective function $E(\twist^t)$ can be efficiently optimized using a Gauss-Newton solver.

\begin{figure}[t]
    \centering
    \includegraphics[width=\linewidth]{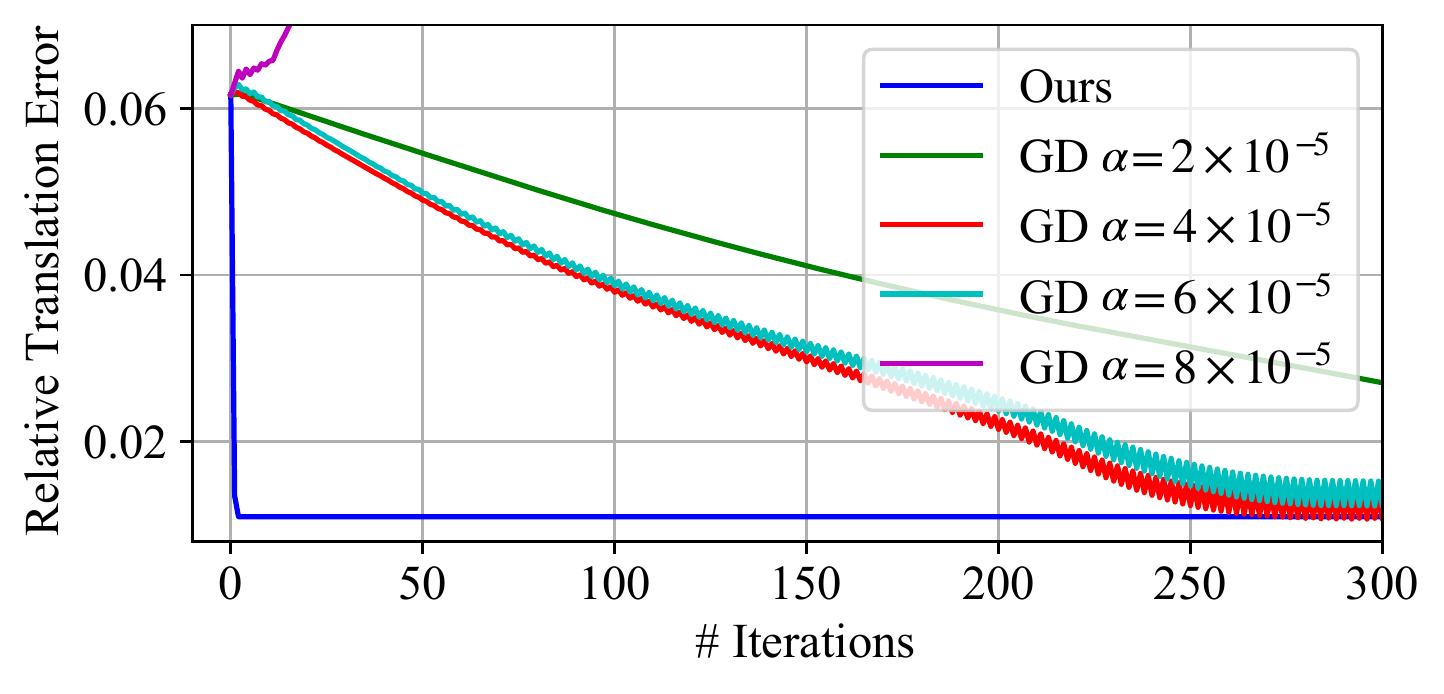}
    \caption{Comparison of converge between different optimization strategies for camera tracking. $\alpha$ is the step size. `Ours' represents the approximate gradient we used in our camera tracking.}
    \label{fig:iter}
\end{figure}

\parahead{SDF Term $E_{\mathrm{sdf}}(\twist^t)$} 
The goal of our SDF term is to perform frame-to-model alignment of the point measurements $\mathcal{P}^t$ to the on-surface geometry decoded by $\mathcal{V}$. 
Different from traditional point-to-plane Iterative Closest Point~(ICP) methods with projective association, as our map representation is fully backboned by implicit functions, we choose to minimize the signed distance value of each point in $\mathcal{P}^t$ when transformed by the optimized camera pose.
Thus we design the objective function as:  
\begin{equation}
\label{equ:sdf}
\begin{aligned}
    E_{\mathrm{sdf}}(\twist^t) &= \sum_{\bm{p}^t \in \mathcal{P}^t} \rho \left( r \left( G(\twist^t, \bm{p}^t) \right)  \right), \\ 
    G(\twist^t, \bm{p}^t) &= \mathbf{T}^{t-1} T(\twist^t) \bm{p}^t, \quad r(\x) =  \frac{ \decmu ( \x, \bm{l}_{m(\x)} ) }{ \decstd ( \x, \bm{l}_{m(\x)} ) },
\end{aligned}
\end{equation}
where $\rho(\cdot)$ is the Huber robust function, $\{\decmu,\decstd\}$ represents the probabilistic signed distance function decoded by the latent vector $\bm{l}_m$ from the point measurements.
Note that $m(\x)$ is a discrete function which may lead to discontinuities across \PV boundaries, but we remark that the final term can be effectively smoothed via the summation of all the points in $\mathcal{P}^t$.

One important step to optimize the SDF term is the computation of $r(\cdot)$'s gradient with respect to $\twist^t$, i.e. $\frac{\partial r}{\partial \twist^t} $.
Note that $\decstd$ and $\decmu$ are the output for decoder $\decoder$, so $r(\cdot)$ is an highly non-linear composite function of decoder $\decoder$, which will lead to poor local approximation.
We instead propose to treat $\decstd$ to be constant during the local linearization, which will efficiently improve convergence during the optimization.
Another benefit is that we can control the influence of $\decstd$ up to the zeroth-order approximation, achieving robust tracking performance even when $\decstd$ prediction is wrong.
Specifically, the approximate gradient is computed by:
\begin{equation}
\label{equ:tracking-gradient}
    \frac{\partial r}{\partial \twist^t} = \frac{1}{\decstd} \frac{\partial \decmu (\cdot, \bm{l}_{m(\x)})}{\partial \bm{x}} (\mathbf{R}^{t-1})^\top ( T(\twist^t) \bm{p}^t )^\odot,
\end{equation}
where $\mathbf{R}^{t-1}$ is the rotation part of $\mathbf{T}^{t-1}$, $\bm{p}^\odot := [\mathbf{I}_{3}, -\bm{p}^{\wedge }]$ and $\frac{\partial \decmu}{\partial \x}$ can be efficiently computed using back-propagation through the decoder network.
As an empirical proof of \cref{equ:tracking-gradient}, we show in \cref{fig:iter} the comparison of convergence property between our method and the gradient decent (GD) method which performs the update step as $\twist^t \leftarrow \twist^t - \alpha \frac{\partial \sum \mathcal{L}_m}{\partial \twist^t}$ using the \emph{precise} gradient.
Our approximation reaches convergence within a few iterations while achieving comparable estimation accuracy compared to GD, demonstrating the effectiveness of our formulation.

\begin{figure}[t]
    \centering
    \includegraphics[width=\linewidth]{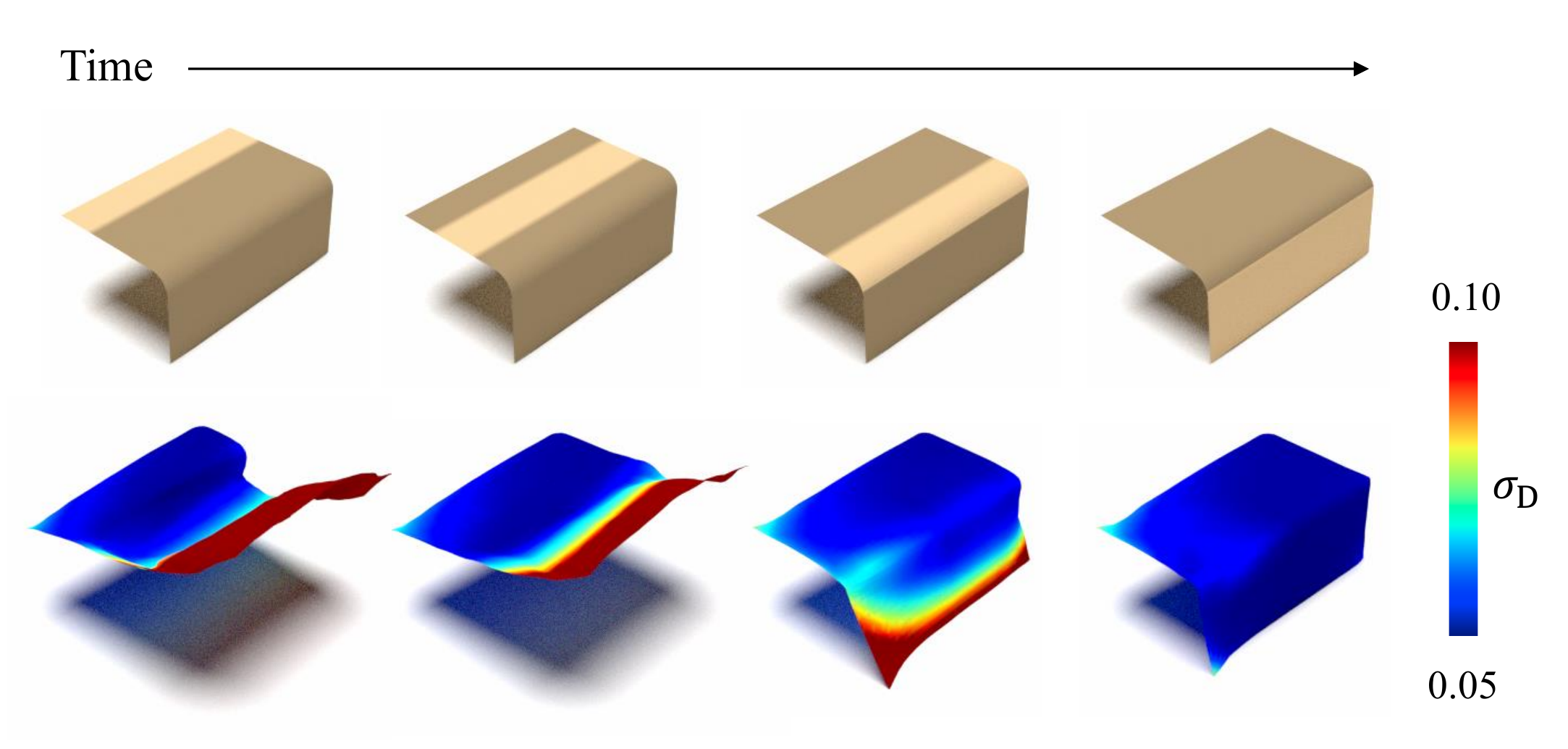}
    \caption{A tiny example demonstrating the effect of incremental integration. The top row shows ground-truth underlying geometry and at each time step, we sample point from the region with lighter color and fuse it into our \PV using \cref{lv:update}. The bottom row shows corresponding geometry after each frame is integrated. The color of the mesh is coded by $\decstd$, reflecting the uncertainty on the surface.}
    \label{fig:integration}
\end{figure}

\parahead{Intensity Term $ E_{\mathrm{int}}(\twist^t)$} We also add a photometric error term between the corresponding intensity data (called intensity term) defined as:
\begin{equation}
    E_{\mathrm{int}}(\twist^t) = \sum_{\mathbf{u} \in \mathbf{\Omega} } \left( \mathcal{I}^t [ \mathbf{u} ] - \mathcal{I}^{t-1} [ \pi( T(\twist^t) \pi' (\mathbf{u}, \mathcal{D}^{t} [\mathbf{u}]) )] \right)^2,
\end{equation}
where $\mathbf{\Omega}$ is the image domain. This intensity term takes effect when the SDF term fails in areas with fewer geometric details such as wall or floor.

\subsection{Surface Mapping}
\label{subsec:method:integration}

After the camera pose of RGB-D observation $\mathcal{O}^t$ is estimated, we need to update the mapping from observation $\mathcal{O}^t$ based on the deep implicit representation, by fusing new scene geometry from new observations with noise, which is also referred to as geometry integration.
Besides, we also describe an optional mesh extraction step that extracts the on-surface watertight mesh with textures for visualization.

\parahead{Geometry Integration}
Instead of updating SDF as previous works~\cite{DBLP:conf/siggraph/CurlessL96}, we propose to update our deep implicit representation in the domain of \emph{latent vectors}.
We perform the geometry integration by updating the geometry latent vector $\bm{l}_m$ with the observation latent vector $\bm{l}_m^t$ encoded by the point measurements $\mathcal{P}^t$. More specifically, we first transform $\mathcal{P}^t$ according to $\mathbf{T}^t$ and then estimate the normal of each point measurement, obtaining $\mathcal{X}^t = \{(\x_i, \normal_i)\}$. 
In each PLIVox $\bm{v}_m$, we calculate the point measurements $\mathcal{Y}_m^t \subset \mathcal{X}^t$ located in such PLIVox and compute the observation latent vector using $\bm{l}_m^t = \frac{1}{w_m^t} \sum_{(\y, \normal)\in \mathcal{Y}^t_m} \encoder(\y, \normal)$. The geometry latent vector $\bm{l}_m$ within such PLIVox $\bm{v}_m$ is then updated as:
\begin{equation}
\label{lv:update}
    \bm{l}_m \leftarrow \frac{\bm{l}_m w_m + \bm{l}_m^t w_m^t}{w_m + w_m^t}, \quad w_m \leftarrow w_m + w_m^t,
\end{equation}
where the weight $w_m^t$ is set to the number of points within the \PV as $|\mathcal{Y}^t_m|$. In this way, our geometry integration is much more efficient than the previous approaches, since we only need to update the latent vectors within a PLIVox but not the SDFs of massive individual voxels as in VoxelHashing~\cite{DBLP:journals/tog/NiessnerZIS13}. 
The observation latent vector from the forward pass of $\encoder$ is also very efficient to compute, which makes our geometry integration fast enough during the online surface mapping without latent vector initialization and optimization used in previous auto-decoder methods~\cite{chabra2020deepls,park2019deepsdf}.
\cref{fig:integration} shows a tiny example demonstrating the effect of incremental geometry integration using our method.

In the meantime, thanks to the robust geometry recovered from the learned deep priors, we do not need to perform integration for every frame. 
Instead, we choose to integrate sparse frames sampled from every $N$ incoming frames, which improve system efficiency largely while still maintaining tracking accuracy.

\parahead{Mesh Extraction}
Optionally we can extract the triangle mesh of the scene surface for visualization purposes. 
Given a desired resolution during the extraction, we divide each \PV into equally-spaced volumetric grids and query the SDFs for each grid with the decoder $\decoder$ using the \PV's latent vector. 
Then the final on-surface mesh can be extracted with Marching Cubes~\cite{lorensen1987marching}. 
Here, to maintain the {continuity} across PLIVox boundaries, we double each \PV's domain such that the volumetric grids between neighboring \PVs overlap with each other. The final SDF of each volumetric grid is trilinearly interpolated with the SDFs decoded from the overlapping \PVs~\cite{jiang2020local}. 
For textures, we simply assign the vertices on the extracted mesh with texture colors averaged from the nearest point measurements back-projected from multiple observations. 

\section{Experiments}
\label{sec:exp}

\subsection{System Implementation}
To train an effective encoder-decoder neural network for descriptive deep priors, we build up two training datasets ($\mathcal{S}$ and $\mathcal{D}$ for encoder and decoder, respectively, \cf \cref{subsec:method:plivox}), on ShapeNet dataset~\cite{chang2015shapenet} which contains a large variety of 3D shapes with rich local geometry details. 
We employ the 6 categories of the 3D shapes from the ShapeNet dataset, \ie bookshelf, display, sofa, chair, lamp, and table, and for each category, we sample 100 3D shapes.
Each shape is then divided into equally-spaced \PVs.
For each \PV $\bm{v}_m$, we randomly sample $n_d=4096$ $(\y_i , s^{i}_{gt})$ tuples for $\mathcal{D}_m$, and the count of $(\y_i,\normal_i)$ tuple $n_s$ for each $\mathcal{S}_m$ is randomly chosen from 16 to 128. 
We further augment $\mathcal{D}_m$ and $\mathcal{S}_m$ by adding random jitter for positions $\y_i$ and perturbation for normal direction $\normal_i$ to train the encoder-decoder neural network with enough robustness against depth observation noise.

We set the length of the latent vector as $L=29$ in both encoder $\encoder$ and decoder $\decoder$ sub-networks.
The encoder $\encoder$ contains 5 fully connected (FC) layers, with the layer sizes set as (6,32,64,256,29). The
decoder contains 6 FC layers with the layer sizes set as (32,128,128,128,128,2). We choose to use the Adam optimizer for training the encoder-ecoder network with an initial learning rate set as $10^{-3}$. For the regularization on the loss function $\mathcal{L}$, we set $\delta=10^{-2}$. 

We manage the \PVs using a similar mechanism as ~\cite{DBLP:journals/tog/NiessnerZIS13} and allocate \PVs only when there are enough point measurements gathered (16 in our experiments). Our DI-Fusion system is implemented using PyTorch framework~\cite{pytorch}: The jacobian matrix of our tracking term~\cref{equ:tracking-gradient} can be efficiently computed with its auto-differentiation.

\begin{table}[!t]
{
\small
\centering
\setlength{\tabcolsep}{8.0pt}
\caption{Comparison of ATE on ICL-NUIM~\cite{handa:etal:ICRA2014} benchmark (measured in centimeters).}
\label{tbl:icl-pose}
\begin{tabularx}{\linewidth}{X|cccc} 
\toprule
& lr kt0        & lr kt1        & lr kt2        & lr kt3          \\ 
\midrule
DVO-SLAM~\cite{DBLP:conf/iros/KerlSC13}      & 10.4           & 2.9           & 19.1           & 15.2            \\
Surfel Tracking~\cite{keller2013real} & \textbf{1.7} & \textbf{1.0}  & 2.2           & 43.2            \\
TSDF Tracking~\cite{prisacariu2017infinitam}   & 4.5           & 2.1           & \textbf{1.3}  & \textbf{12.5}  \\ 
\midrule
Ours (w/o Prob) & 1.3           & 1.8           & 2.6           & 15.2            \\
Ours (max)      & 1.4           & 2.0           & \textbf{2.4} & 7.1             \\
Ours     & \textbf{1.1}  & \textbf{1.4} & 2.6           & \textbf{6.2}    \\
\bottomrule
\end{tabularx}
}
\end{table}

\begin{table}[!t]
{
\small
\centering
\setlength{\tabcolsep}{8.0pt}
\caption{Comparison of surface error on ICL-NUIM~\cite{handa:etal:ICRA2014} benchmark (measured in centimeters). $\decstd$ is thresholded to 0.06.}
\label{tbl:icl-surface}
\begin{tabularx}{\linewidth}{X|cccc} 
\toprule
& lr kt0        & lr kt1        & lr kt2        & lr kt3          \\ 
\midrule
DVO-SLAM~\cite{DBLP:conf/iros/KerlSC13}      & 3.2          & 6.1          & 11.9          & \textbf{5.3}           \\
Surfel Tracking~\cite{keller2013real} & 1.1          & \textbf{0.7} & 1.0          & 22.5           \\
TSDF Tracking~\cite{prisacariu2017infinitam}   & \textbf{0.6} & 1.0          & \textbf{0.8} & 11.7  \\ 
\midrule
Ours (w/o Prob) & 0.7          & 2.0          & 1.2          & 4.8            \\
Ours (max)      & 0.8          & 1.8          & 1.2          & 4.8            \\
Ours     & \textbf{0.6} & \textbf{1.5} & \textbf{1.1} & \textbf{4.5}   \\
\bottomrule
\end{tabularx}
}
\vspace{-3mm}
\end{table}

\subsection{Quantitative Evaluations}
\label{sec:quantity}

\parahead{Dataset and Metrics}
In this subsection, we demonstrate the effectiveness of DI-Fusion and its core component, \ie the \PV representation, on ICL-NUIM dataset~\cite{handa:etal:ICRA2014}, which contains both the ground-truth camera trajectory and the 3D scene geometry to evaluate the accuracy of depth fusion approaches.
We adopt the Absolute Trajectory Error (ATE) metric for camera pose estimation and the surface error~\cite{DBLP:conf/rss/WhelanLSGD15} for 3D surface quality evaluation.

\parahead{Baselines}
Since our algorithm does not contain loop closure component, for a fair comparison we choose to compare with previous depth fusion approaches using TSDF or surfel representation with loop closure turned off.
For TSDF-based camera tracking, we adopt the frame-to-model ICP tracker implemented in~\cite{prisacariu2017infinitam}, denoted as `TSDF tracking'. 
For surfel-based camera tracking, we adopt the method in~\cite{DBLP:conf/rss/WhelanLSGD15}, denoted as `surfel tracking'.
Additionally, we compare to DVO-SLAM~\cite{DBLP:conf/iros/KerlSC13}, which performs frame-to-frame camera tracking directly \emph{without} the aid of any 3D scene representations.
Note that both the depth and intensity information are used in all the baselines.

\begin{figure*}[htbp]
    \centering
    \includegraphics[width=\linewidth]{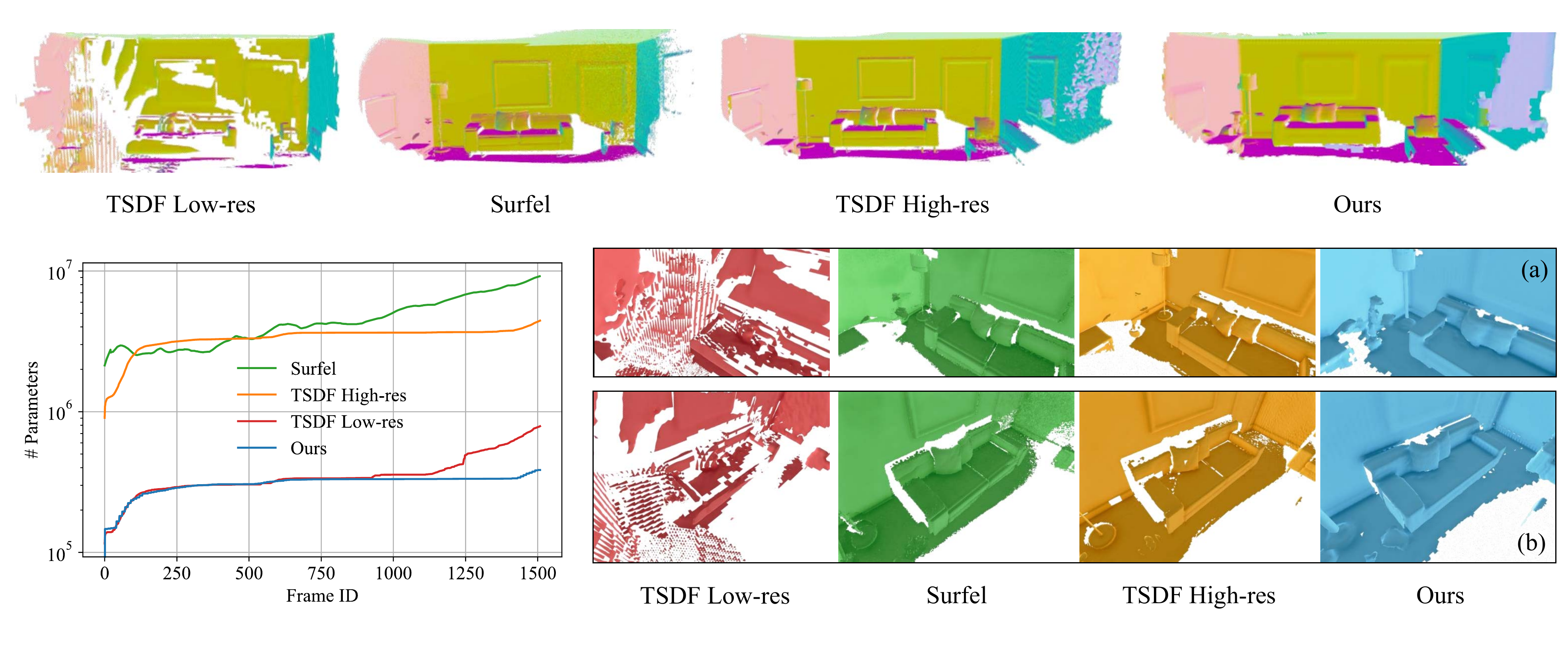}
    \vspace{-2em}
    \caption{Qualitative comparisons and memory analysis on the \texttt{lr\_kt0} sequence of ICL-NUIM~\cite{handa:etal:ICRA2014} dataset. On the top row we show a global view of the reconstructed 3D scene, where the colors represent the normal directions. Close-up looks at the details are visualized at the bottom-right inset (a) (b). We show the evolution of the number of parameters used in each approach at the bottom-left inset.
    }
    \label{fig:icl-compare}
\end{figure*}

\parahead{Results}
As shown in \cref{tbl:icl-pose}, our approach achieves lower or comparable ATE compared to the baselines.
This is mainly due to two reasons:
(1) The deep priors learned in our \PV provide a continuous underlying surface prediction than TSDF or surfel, which provides smoother gradient estimation for camera tracking, and
(2) the uncertainty estimated by our network also effectively filters out noisy RGB-D observations, thus being less prone to large tracking failure.
We find it challenging for our method to fit extremely small structures with the current voxel size ($a=10\mathrm{cm}$) primarily due to the size limitation of the prior space.
Nevertheless, our method is found to be stable and much more memory-efficient than our counterparts.
The surface error comparison shown in~\cref{tbl:icl-surface} verifies the analysis above. 
Thanks to the learned deep priors provided by our \PV representation, our approach reaches state-of-the-art surface error compared to TSDF or surfel-based representations, with much fewer parameters required as shown in the bottom-left inset of \cref{fig:icl-compare}.

\begin{figure*}[htbp]
    \centering
    \includegraphics[width=\linewidth]{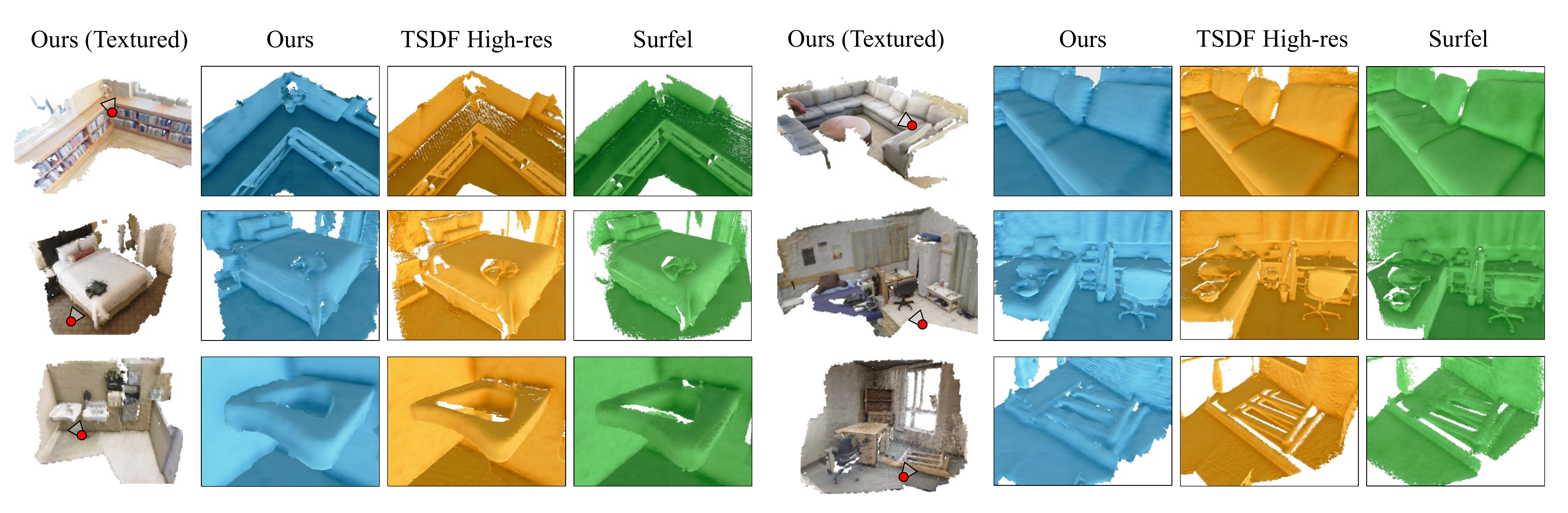}
    \vspace{-1.5em}
    \caption{Qualitative comparisons on ScanNet~\cite{dai2017scannet} dataset. The \nth{1}  column of every scene shows a global view of our reconstruction with textures applied while detailed close-up comparisons with other baselines are shown in other columns. The close-up views' camera positions are plotted on the textured reconstruction.}
    \label{fig:scannet-compare}
\end{figure*}

\begin{figure}[htbp]
    \centering
    \vspace{-1.5em}
    \includegraphics[width=\linewidth]{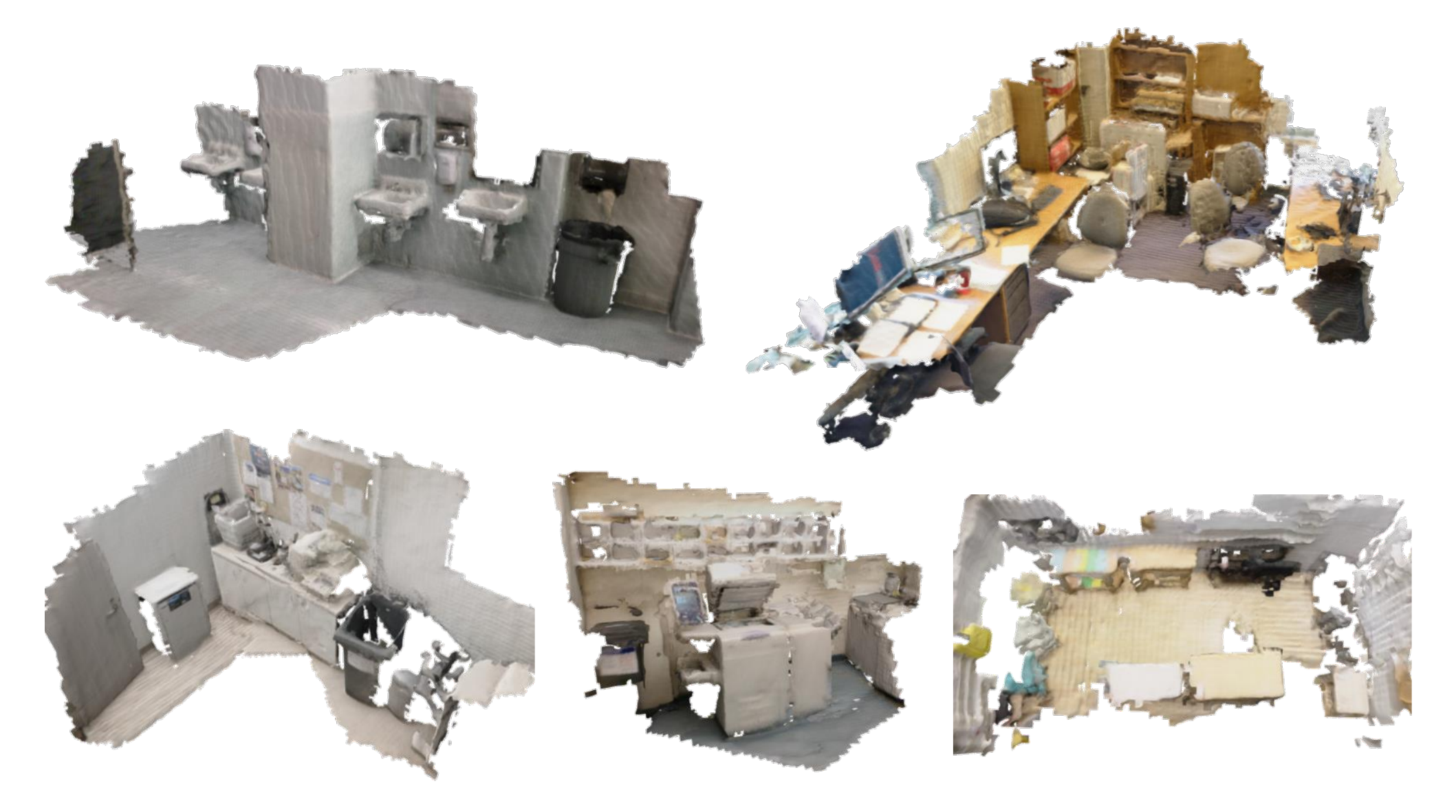}
    \caption{Other qualitative textured results on ScanNet~\cite{dai2017scannet} reconstructed online with our method.}
    \label{fig:scannet-others}
    \vspace{-1.5em}
\end{figure}

\vspace{-0.3em}
\subsection{Qualitative Results}
\label{sec:quality}
\vspace{-0.5em}

We compare the visual quality of the reconstructed 3D surfaces between our approach and the other two camera tracking approaches (TSDF tracking and surfel tracking). 
We implement TSDF tracking with two configurations: 
(1) `TSDF Low-res' (low resolution) which allocates larger voxels, so that the number of parameters used to represent the geometry roughly matches the storage cost of our approach; 
(2) `TSDF High-res' (high resolution, also used in the quantitative comparisons in \cref{sec:quantity}) with a much smaller voxel size for a higher quality 3D surface reconstruction. 
\cref{fig:icl-compare} shows the visual effect comparison on a scan in \cite{handa:etal:ICRA2014} among different tracking approaches. 
Our method can achieve a more complete 3D surface reconstruction while using $91.3\%$ and $95.8\%$ fewer parameters than `TSDF High-res' and surfel tracking respectively to represent the entire scene. 
If the same amount of memory is used as ours, both the camera tracking and the reconstruction quality would be severely hampered for TSDF approaches as shown in `TSDF Low-res' results.

Additionally, we compare the qualitative results between the different approaches on ScanNet~\cite{dai2017scannet}, which contains large-scale real-world 3D indoor scans captured with a commodity RGB-D camera. 
As shown in \cref{fig:scannet-compare}, our approach can output more preferable 3D reconstructions than the other two baselines due to our \PV's learned encoding of scene priors such that the 3D surface can be accurately recovered at the position even without enough observations. 
Note that we only use the trained weights on ShapeNet dataset for the encoder-decoder network without any subsequent fine-tuning on ScanNet dataset, showing that our representation has a good generalization performance across different scene types.   
Fig.~\ref{fig:scannet-others} shows more illustrations of our approach on ScanNet dataset.
\ifarxiv
Please visit \videourl{} for further demonstrations.
\else
Please refer to our supplementary video for further demonstrations.
\fi

\begin{figure}[t]
    \centering
    \vspace{-1.5em}
    \includegraphics[width=\linewidth]{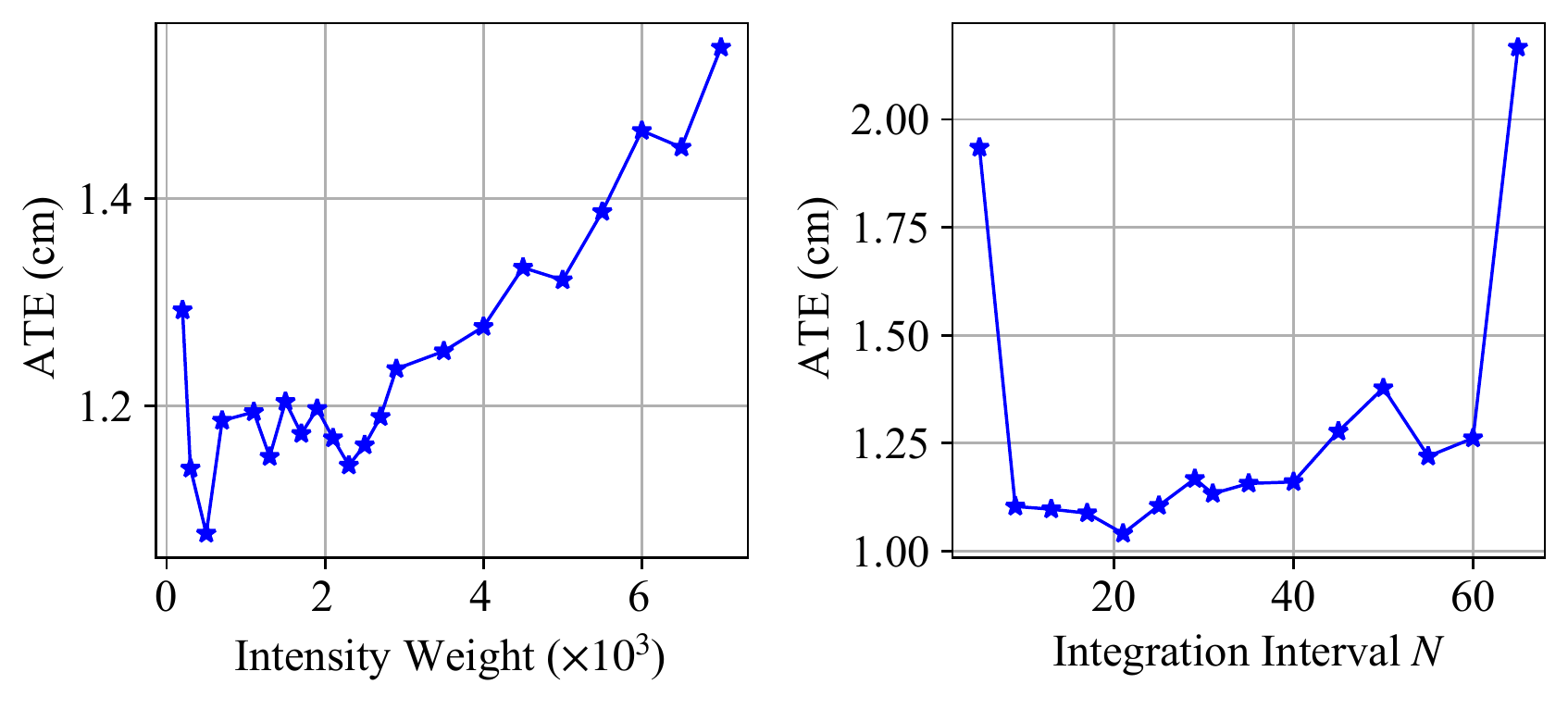}
    \caption{Influence of different parameters on ATE benchmark of ICL-NUIM~\cite{handa:etal:ICRA2014}.}
    \label{fig:parameter}
    \vspace{-1.5em}
\end{figure}

\subsection{System Ablations and Analysis}
\label{sec:abstudy}

\parahead{Probabilistic Modeling} 
To evaluate the effect of using probabilistic SDF for tracking, we ignore the decoder branch outputting $\decstd$ and perform camera tracking by assuming the standard deviation as a constant 1.0 (denoted as `Ours w/o Prob'). 
As demonstrated by the consistent worse results than `Ours' for both camera trajectory estimation (\cref{tbl:icl-pose}) and surface error (\cref{tbl:icl-surface}), we remark that the way we encode the scene priors in a \emph{probabilistic} manner is beneficial because erroneous fitting is properly down-weighted.

\parahead{Intensity Term} The intensity term used in our camera tracking (Sec.~\ref{subsec:method:tracking}) also influences the camera pose estimation accuracy. 
Here we modify our camera tracking with different weights of the intensity term in the objective function (Eq.~\ref{equ:opt}). 
The ATE curve on the ICL-NUIM sequence with respect to different intensity weights is illustrated in Fig.~\ref{fig:parameter} (left), showing the complementary effect between the two terms we used:
Pure SDF tracking (\eg $w=0$) tends to fail in places with few geometric features and too large intensity weights (\eg $w > 4000$) will ignore the reconstructed scene representation, causing increased drift.
In practice, setting the weight of the intensity term between $500$ and $2000$ leads to the best result, in which case the absolute scale of the SDF term $E_\mathrm{sdf}$ and intensity term $E_\mathrm{int}$ for accurate camera pose estimation is roughly balanced.

\begin{figure}[t]
    \centering
    \includegraphics[width=\linewidth]{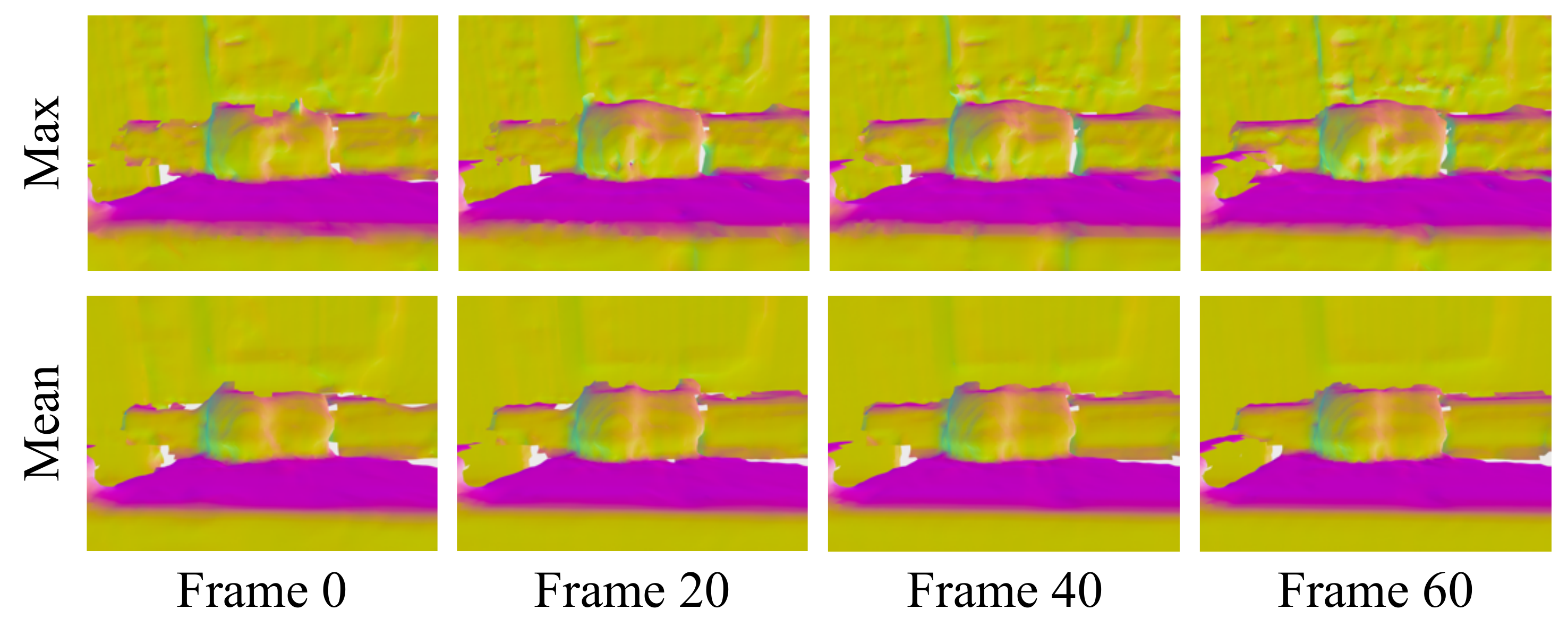}
    \caption{Comparison with alternative integration method using the max operator. Each column shows the actual frames we perform the integration.}
    \label{fig:max-mean}
    \vspace{-1em}
\end{figure}

\parahead{Latent Vector Update}
One alternative way for the geometry integration is to update the latent vector using the max operator as in~\cite{qi2017pointnet} instead of what we propose in \cref{lv:update}, \ie $\bm{l}_m \leftarrow \max(\bm{l}_m,\bm{l}_m^t)$ and we call this baseline `Ours (max)'.
\cref{tbl:icl-pose,tbl:icl-surface} show the average ATE and surface error of this approach, which are consistently worse than using the mean operator. 
One possible reason would be that the max operator is sensitive to sensor noise, thus leading to spurious reconstructions, while our averaging update could provide smoother reconstructions against sensor noise. 
\cref{fig:max-mean} shows a comparison between the two integration schemes.

\parahead{Geometry Integration} The way we perform the geometry integration at sparse frames sampled from every $N$ frames would also influence the surface mapping quality. 
Fig.~\ref{fig:parameter} (right) shows the average ATE curve for the camera pose estimation along with different frame integration intervals. 
The average ATE becomes larger along with the increase of integration interval. In contrast, too frequent integration will introduce excessive sensor noise, which confuses the network and deteriorates the mapping quality.

\begin{figure}[t]
   \begin{center}
      \includegraphics[width=0.95\linewidth]{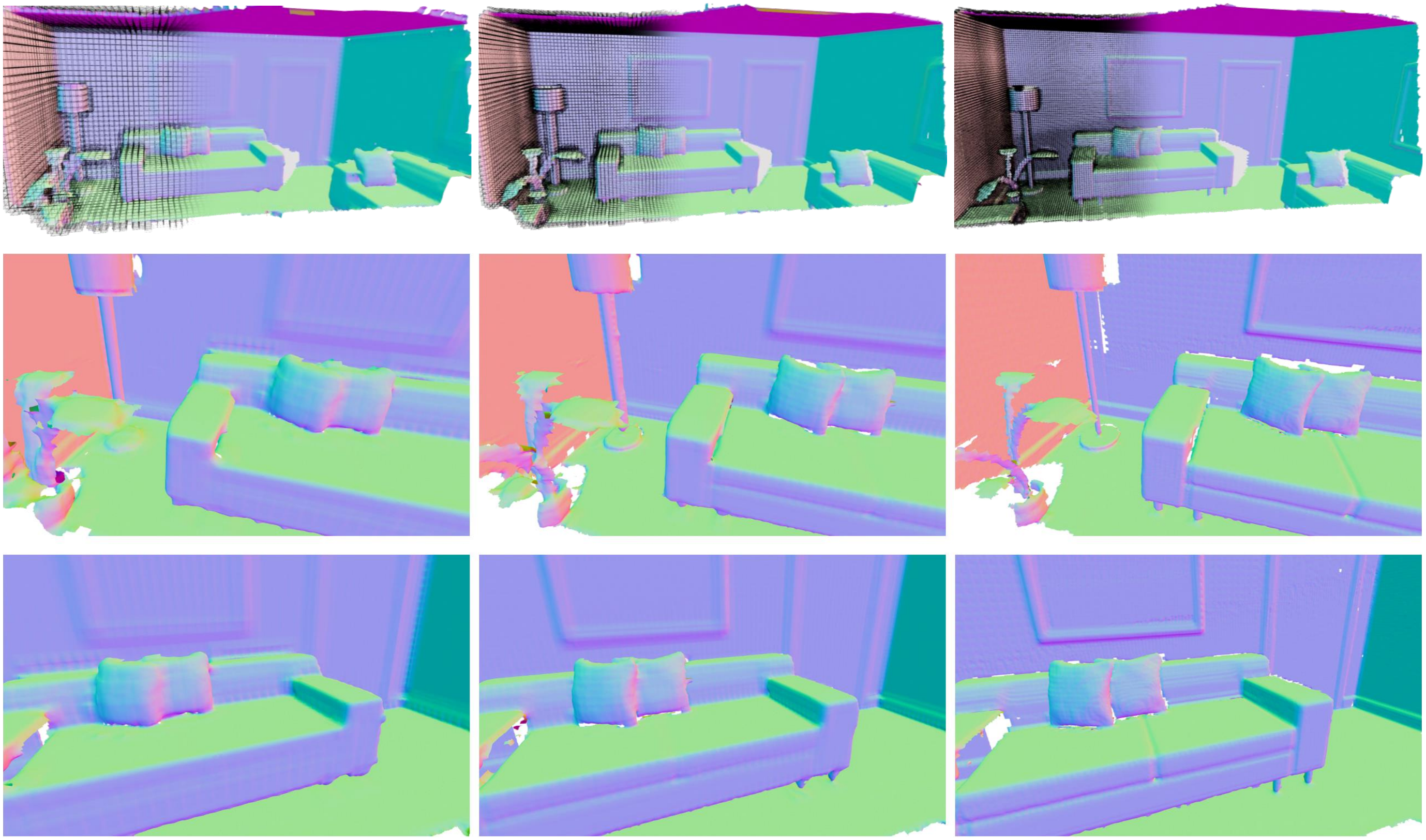}
   \end{center}
   \vspace{-1.3em}
   \setlength{\tabcolsep}{24.0pt}
   {\footnotesize
   \begin{tabular}{ccc}
    $a=8\mathrm{cm}$ & $a=5\mathrm{cm}$ & $a=3\mathrm{cm}$
   \end{tabular}}
   \caption{Reconstruction results with different \PV sizes $a$.}
\label{fig:vsize}
\vspace{-0.8em}
\end{figure}

\parahead{Voxel Size}
We show one reconstructed scene using different \PV sizes in \cref{fig:vsize}.
By shrinking the \PV size $a$, the reconstruction quality is boosted with more desirable details.
However, the running time and the memory requirement will also increase accordingly.
An engineered implementation with a double-layer \PV representation can be employed where large voxels are used for real-time tracking and small voxels are for final mapping. 

\begin{figure}[t]
    \begin{center}
       \includegraphics[width=\linewidth]{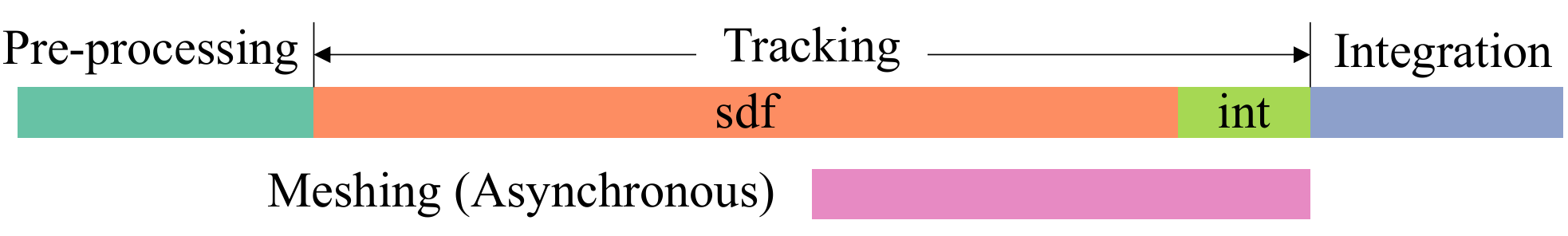}
    \end{center}
    \vspace{-1em}
    \caption{Timing analysis. The bar length represents the relative time cost of each component in our system. `sdf' and `int' mean the computation related to $E_{\mathrm{sdf}}$ and $E_{\mathrm{int}}$, respectively.}
 \label{fig:timing}
\vspace{-1.2em} 
\end{figure}

\parahead{Timing}
Overall, our system can run online at 10-12Hz on most modern GPU platforms we test.
The cost of each component in our system is reported in \cref{fig:timing}.
Further code optimizations are possible with parallelization, caching, etc., which we remark as future works.
The post-processing time for texturing is $\sim$50ms per integrated frame, with the nearest neighbor search accelerated by a k-d tree.

\section{Conclusions}

\paragraph{Limitations.}
Our approach has three main limitations: (1) The learned prior could not provide an omnipotent fitting of all possible local geometries especially in the case of over-complicated objects. (2) Each \PV is independent and the relationships between neighboring \PVs are not considered; therefore the spatial continuity of the reconstructed scene is not guaranteed. (3) No loop closure component has been incorporated to enforce global consistency.
We hope to investigate deeply into both the limitations in the future.

In this paper, we present DI-Fusion, which performs online implicit 3D reconstruction with deep priors.
With a novel \PV representation, our approach effectively incorporates scene priors of both geometry and uncertainty into joint camera tracking and surface mapping, achieving more accurate camera pose estimation and higher-quality 3D reconstruction. 
We hope that our method can inspire more future efforts for advanced online 3D reconstruction. 

\vspace{0.5em}

\begin{spacing}{0.8}
{\footnotesize
\noindent\textbf{Acknowledgements.}
We thank anonymous reviewers for the valuable discussions.
This work was supported by the Natural Science Foundation of China (Project No. 61521002), the Joint NSFC-DFG Research Program (Project No. 61761136018), Research Grant of Tsinghua-Tencent Joint Laboratory for Internet Innovation Technology, and grants from the China Postdoctoral Science Foundation (Grant No.: 2019M660646).}
\end{spacing}

{\small
\bibliographystyle{ieee_fullname}
\bibliography{egbib}

\begin{thebibliography}{10}\itemsep=-1pt

\bibitem{barfoot2017state}
Timothy~D Barfoot.
\newblock {\em State estimation for robotics}.
\newblock Cambridge University Press, 2017.

\bibitem{DBLP:conf/cvpr/BloeschCCLD18}
Michael Bloesch, Jan Czarnowski, Ronald Clark, Stefan Leutenegger, and
  Andrew~J. Davison.
\newblock Codeslam - learning a compact, optimisable representation for dense
  visual {SLAM}.
\newblock In {\em IEEE CVPR}, pages 2560--2568, 2018.

\bibitem{Cao:2018:RHT:3278329.3182157}
Yan{-}Pei Cao, Leif Kobbelt, and Shi{-}Min Hu.
\newblock Real-time high-accuracy three-dimensional reconstruction with
  consumer {RGB-D} cameras.
\newblock {\em {ACM} Trans. Graph.}, 37(5):171:1--171:16, 2018.

\bibitem{chabra2020deepls}
Rohan Chabra, Jan~Eric Lenssen, Eddy Ilg, Tanner Schmidt, Julian Straub, Steven
  Lovegrove, and Richard~A. Newcombe.
\newblock Deep local shapes: Learning local {SDF} priors for detailed 3d
  reconstruction.
\newblock In Andrea Vedaldi, Horst Bischof, Thomas Brox, and Jan{-}Michael
  Frahm, editors, {\em ECCV}, volume 12374, pages 608--625, 2020.

\bibitem{chang2015shapenet}
Angel~X Chang, Thomas Funkhouser, Leonidas Guibas, Pat Hanrahan, Qixing Huang,
  Zimo Li, Silvio Savarese, Manolis Savva, Shuran Song, Hao Su, et~al.
\newblock Shapenet: An information-rich 3d model repository.
\newblock {\em arXiv preprint arXiv:1512.03012}, 2015.

\bibitem{DBLP:journals/tog/ChenBI13}
Jiawen Chen, Dennis Bautembach, and Shahram Izadi.
\newblock Scalable real-time volumetric surface reconstruction.
\newblock {\em {ACM} Trans. Graph.}, 32(4):113:1--113:16, 2013.

\bibitem{chen2019bae}
Zhiqin Chen, Kangxue Yin, Matthew Fisher, Siddhartha Chaudhuri, and Hao Zhang.
\newblock Bae-net: Branched autoencoder for shape co-segmentation.
\newblock In {\em IEEE ICCV}, pages 8490--8499, 2019.

\bibitem{chen2019imnet}
Zhiqin Chen and Hao Zhang.
\newblock Learning implicit fields for generative shape modeling.
\newblock In {\em IEEE CVPR}, pages 5939--5948, 2019.

\bibitem{choi2015robust}
Sungjoon Choi, Qian-Yi Zhou, and Vladlen Koltun.
\newblock Robust reconstruction of indoor scenes.
\newblock In {\em IEEE CVPR}, pages 5556--5565, 2015.

\bibitem{DBLP:conf/siggraph/CurlessL96}
Brian Curless and Marc Levoy.
\newblock A volumetric method for building complex models from range images.
\newblock In {\em {ACM SIGGRAPH}}, pages 303--312, 1996.

\bibitem{dai2017scannet}
Angela Dai, Angel~X. Chang, Manolis Savva, Maciej Halber, Thomas Funkhouser,
  and Matthias Nie{\ss}ner.
\newblock Scannet: Richly-annotated 3d reconstructions of indoor scenes.
\newblock In {\em IEEE CVPR}, 2017.

\bibitem{dai2017bundlefusion}
Angela Dai, Matthias Nie{\ss}ner, Michael Zollh{\"{o}}fer, Shahram Izadi, and
  Christian Theobalt.
\newblock Bundlefusion: Real-time globally consistent 3d reconstruction using
  on-the-fly surface reintegration.
\newblock {\em {ACM} Trans. Graph.}, 36(3):24:1--24:18, 2017.

\bibitem{DBLP:conf/cvpr/DaiRBRSN18}
Angela Dai, Daniel Ritchie, Martin Bokeloh, Scott Reed, J{\"{u}}rgen Sturm, and
  Matthias Nie{\ss}ner.
\newblock Scancomplete: Large-scale scene completion and semantic segmentation
  for 3d scans.
\newblock In {\em IEEE CVPR}, pages 4578--4587, 2018.

\bibitem{DBLP:journals/pami/DavisonRMS07}
A.~J. {Davison}, I.~D. {Reid}, N.~D. {Molton}, and O. {Stasse}.
\newblock Monoslam: Real-time single camera slam.
\newblock {\em IEEE TPAMI}, 29(6):1052--1067, 2007.

\bibitem{dong2018psdf}
Wei Dong, Qiuyuan Wang, Xin Wang, and Hongbin Zha.
\newblock Psdf fusion: Probabilistic signed distance function for on-the-fly 3d
  data fusion and scene reconstruction.
\newblock In {\em ECCV}, pages 701--717, 2018.

\bibitem{DBLP:journals/pami/EngelKC18}
J. {Engel}, V. {Koltun}, and D. {Cremers}.
\newblock Direct sparse odometry.
\newblock {\em IEEE TPAMI}, 40(3):611--625, 2018.

\bibitem{DBLP:conf/eccv/EngelSC14}
Jakob Engel, Thomas Sch{\"{o}}ps, and Daniel Cremers.
\newblock {LSD-SLAM:} large-scale direct monocular {SLAM}.
\newblock In {\em ECCV}, pages 834--849, 2014.

\bibitem{garnelo2018cnp}
Marta Garnelo, Dan Rosenbaum, Christopher Maddison, Tiago Ramalho, David
  Saxton, Murray Shanahan, Yee~Whye Teh, Danilo~Jimenez Rezende, and S.~M.~Ali
  Eslami.
\newblock Conditional neural processes.
\newblock In Jennifer~G. Dy and Andreas Krause, editors, {\em ICML}, volume~80,
  pages 1690--1699, 2018.

\bibitem{Deeptrack2017}
M. {Garon} and J. {Lalonde}.
\newblock Deep 6-dof tracking.
\newblock {\em IEEE TVCG}, 23(11):2410--2418, 2017.

\bibitem{handa:etal:ICRA2014}
A. Handa, T. Whelan, J.B. McDonald, and A.J. Davison.
\newblock A benchmark for {RGB-D} visual odometry, {3D} reconstruction and
  {SLAM}.
\newblock In {\em IEEE ICRA}, Hong Kong, China, May 2014.

\bibitem{huang2020wallnet}
Jiahui Huang, Zheng-Fei Kuang, Fang-Lue Zhang, and Tai-Jiang Mu.
\newblock Wallnet: Reconstructing general room layouts from rgb images.
\newblock {\em Graphical Models}, 111:101076, 2020.

\bibitem{huang2020clustervo}
Jiahui Huang, Sheng Yang, Tai-Jiang Mu, and Shi-Min Hu.
\newblock Clustervo: Clustering moving instances and estimating visual odometry
  for self and surroundings.
\newblock In {\em IEEE CVPR}, pages 2168--2177, 2020.

\bibitem{huang2019clusterslam}
Jiahui Huang, Sheng Yang, Zishuo Zhao, Yu-Kun Lai, and Shi-Min Hu.
\newblock Clusterslam: A slam backend for simultaneous rigid body clustering
  and motion estimation.
\newblock In {\em IEEE ICCV}, pages 5875--5884, 2019.

\bibitem{jiang2020local}
Chiyu Jiang, Avneesh Sud, Ameesh Makadia, Jingwei Huang, Matthias Nie{\ss}ner,
  and Thomas Funkhouser.
\newblock Local implicit grid representations for 3d scenes.
\newblock In {\em IEEE CVPR}, pages 6001--6010, 2020.

\bibitem{DBLP:journals/tvcg/KahlerPRSTM15}
Olaf K{\"{a}}hler, Victor~Adrian Prisacariu, Carl~Yuheng Ren, Xin Sun, Philip
  H.~S. Torr, and David~William Murray.
\newblock Very high frame rate volumetric integration of depth images on mobile
  devices.
\newblock {\em {IEEE} TVCG.}, 21(11):1241--1250, 2015.

\bibitem{keller2013real}
Maik Keller, Damien Lefloch, Martin Lambers, Shahram Izadi, Tim Weyrich, and
  Andreas Kolb.
\newblock Real-time 3d reconstruction in dynamic scenes using point-based
  fusion.
\newblock In {\em 3DV}, pages 1--8. IEEE, 2013.

\bibitem{DBLP:conf/iros/KerlSC13}
Christian Kerl, J{\"{u}}rgen Sturm, and Daniel Cremers.
\newblock Dense visual {SLAM} for {RGB-D} cameras.
\newblock In {\em IEEE IROS}, pages 2100--2106, 2013.

\bibitem{laidlow2019deepfusion}
Tristan Laidlow, Jan Czarnowski, and Stefan Leutenegger.
\newblock Deepfusion: real-time dense 3d reconstruction for monocular slam
  using single-view depth and gradient predictions.
\newblock In {\em IEEE ICRA}, pages 4068--4074, 2019.

\bibitem{lee2019online}
Bhoram Lee, Clark Zhang, Zonghao Huang, and Daniel~D Lee.
\newblock Online continuous mapping using gaussian process implicit surfaces.
\newblock In {\em IEEE ICRA}, pages 6884--6890, 2019.

\bibitem{liu2020dist}
Shaohui Liu, Yinda Zhang, Songyou Peng, Boxin Shi, Marc Pollefeys, and Zhaopeng
  Cui.
\newblock Dist: Rendering deep implicit signed distance function with
  differentiable sphere tracing.
\newblock In {\em IEEE CVPR}, pages 2019--2028, 2020.

\bibitem{lorensen1987marching}
William~E Lorensen and Harvey~E Cline.
\newblock Marching cubes: A high resolution 3d surface construction algorithm.
\newblock {\em ACM siggraph computer graphics}, 21(4):163--169, 1987.

\bibitem{DBLP:journals/ral/MartensPSFS17}
Wolfram Martens, Yannick Poffet, Pablo~Ramon Soria, Robert Fitch, and Salah
  Sukkarieh.
\newblock Geometric priors for gaussian process implicit surfaces.
\newblock {\em {IEEE} Robotics Autom. Lett.}, 2(2):373--380, 2017.

\bibitem{DBLP:conf/3dim/McCormacCBDL18}
John McCormac, Ronald Clark, Michael Bloesch, Andrew~J. Davison, and Stefan
  Leutenegger.
\newblock Fusion++: Volumetric object-level {SLAM}.
\newblock In {\em 3DV}, pages 32--41, 2018.

\bibitem{mescheder2019occupancy}
Lars Mescheder, Michael Oechsle, Michael Niemeyer, Sebastian Nowozin, and
  Andreas Geiger.
\newblock Occupancy networks: Learning 3d reconstruction in function space.
\newblock In {\em IEEE CVPR}, pages 4460--4470, 2019.

\bibitem{mildenhall2020nerf}
Ben Mildenhall, Pratul~P Srinivasan, Matthew Tancik, Jonathan~T Barron, Ravi
  Ramamoorthi, and Ren Ng.
\newblock Nerf: Representing scenes as neural radiance fields for view
  synthesis.
\newblock In {\em ECCV}, pages 405--421. Springer, 2020.

\bibitem{DBLP:journals/trob/Mur-ArtalT17}
R. {Mur-Artal} and J.~D. {Tard\'{o}s}.
\newblock Orb-slam2: An open-source slam system for monocular, stereo, and
  rgb-d cameras.
\newblock {\em IEEE Transactions on Robotics (TRO)}, 33(5):1255--1262, 2017.

\bibitem{newcombe2015dynamicfusion}
Richard~A Newcombe, Dieter Fox, and Steven~M Seitz.
\newblock Dynamicfusion: Reconstruction and tracking of non-rigid scenes in
  real-time.
\newblock In {\em IEEE CVPR}, pages 343--352, 2015.

\bibitem{KinectFusion}
Richard~A. Newcombe, Shahram Izadi, Otmar Hilliges, David Molyneaux, David Kim,
  Andrew~J. Davison, Pushmeet Kohli, Jamie Shotton, Steve Hodges, and Andrew~W.
  Fitzgibbon.
\newblock Kinectfusion: Real-time dense surface mapping and tracking.
\newblock In {\em {IEEE ISMAR}}, pages 127--136, 2011.

\bibitem{DBLP:journals/tog/NiessnerZIS13}
Matthias Nie{\ss}ner, Michael Zollh{\"{o}}fer, Shahram Izadi, and Marc
  Stamminger.
\newblock Real-time 3d reconstruction at scale using voxel hashing.
\newblock {\em {ACM} Trans. Graph.}, 32(6):169:1--169:11, 2013.

\bibitem{park2019deepsdf}
Jeong~Joon Park, Peter Florence, Julian Straub, Richard Newcombe, and Steven
  Lovegrove.
\newblock Deepsdf: Learning continuous signed distance functions for shape
  representation.
\newblock In {\em IEEE CVPR}, pages 165--174, 2019.

\bibitem{pytorch}
Adam Paszke, Sam Gross, Francisco Massa, Adam Lerer, James Bradbury, Gregory
  Chanan, Trevor Killeen, Zeming Lin, Natalia Gimelshein, Luca Antiga, Alban
  Desmaison, Andreas Kopf, Edward Yang, Zachary DeVito, Martin Raison, Alykhan
  Tejani, Sasank Chilamkurthy, Benoit Steiner, Lu Fang, Junjie Bai, and Soumith
  Chintala.
\newblock Pytorch: An imperative style, high-performance deep learning library.
\newblock In H. Wallach, H. Larochelle, A. Beygelzimer, F. d\'Alch\'{e} Buc, E.
  Fox, and R. Garnett, editors, {\em Advances in Neural Information Processing
  Systems 32}, pages 8024--8035. Curran Associates, Inc., 2019.

\bibitem{prisacariu2017infinitam}
Victor~Adrian Prisacariu, Olaf K{\"a}hler, Stuart Golodetz, Michael Sapienza,
  Tommaso Cavallari, Philip~HS Torr, and David~W Murray.
\newblock Infinitam v3: A framework for large-scale 3d reconstruction with loop
  closure.
\newblock {\em arXiv preprint arXiv:1708.00783}, 2017.

\bibitem{qi2017pointnet}
Charles~R Qi, Hao Su, Kaichun Mo, and Leonidas~J Guibas.
\newblock Pointnet: Deep learning on point sets for 3d classification and
  segmentation.
\newblock In {\em IEEE CVPR}, pages 652--660, 2017.

\bibitem{sitzmann2020metasdf}
Vincent Sitzmann, Eric~R. Chan, Richard Tucker, Noah Snavely, and Gordon
  Wetzstein.
\newblock Metasdf: Meta-learning signed distance functions.
\newblock In {\em NeurIPS}, 2020.

\bibitem{sitzmann2020implicit}
Vincent Sitzmann, Julien Martel, Alexander Bergman, David Lindell, and Gordon
  Wetzstein.
\newblock Implicit neural representations with periodic activation functions.
\newblock {\em NeurIPS}, 33, 2020.

\bibitem{stutz2020learning}
David Stutz and Andreas Geiger.
\newblock Learning 3d shape completion under weak supervision.
\newblock {\em International Journal of Computer Vision}, 128(5):1162--1181,
  2020.

\bibitem{Sucar:etal:3DV2020}
Edgar Sucar, Kentaro Wada, and Andrew Davison.
\newblock {NodeSLAM}: Neural object descriptors for multi-view shape
  reconstruction.
\newblock In {\em 3DV}, 2020.

\bibitem{DBLP:conf/cvpr/TatenoTLN17}
Keisuke Tateno, Federico Tombari, Iro Laina, and Nassir Navab.
\newblock {CNN-SLAM:} real-time dense monocular {SLAM} with learned depth
  prediction.
\newblock In {\em IEEE CVPR}, pages 6565--6574, 2017.

\bibitem{weder2020routedfusion}
Silvan Weder, Johannes Schonberger, Marc Pollefeys, and Martin~R Oswald.
\newblock Routedfusion: Learning real-time depth map fusion.
\newblock In {\em IEEE CVPR}, pages 4887--4897, 2020.

\bibitem{DBLP:journals/ijrr/WhelanKJFLM15}
Thomas Whelan, Michael Kaess, Hordur Johannsson, Maurice~F. Fallon, John~J.
  Leonard, and John McDonald.
\newblock Real-time large-scale dense {RGB-D} {SLAM} with volumetric fusion.
\newblock {\em I. J. Robotics Res.}, 34(4-5):598--626, 2015.

\bibitem{DBLP:conf/rss/WhelanLSGD15}
Thomas Whelan, Stefan Leutenegger, Renato~F. Salas{-}Moreno, Ben Glocker, and
  Andrew~J. Davison.
\newblock Elasticfusion: Dense {SLAM} without {A} pose graph.
\newblock In {\em Robotics: Science and Systems}, 2015.

\bibitem{yang2020noise}
Sheng Yang, Beichen Li, Yan-Pei Cao, Hongbo Fu, Yu-Kun Lai, Leif Kobbelt, and
  Shi-Min Hu.
\newblock Noise-resilient reconstruction of panoramas and 3d scenes using
  robot-mounted unsynchronized commodity rgb-d cameras.
\newblock {\em {ACM} Trans. Graph.}, 39(5):1--15, 2020.

\bibitem{DBLP:journals/cvgip/ZengZZL13}
Ming Zeng, Fukai Zhao, Jiaxiang Zheng, and Xinguo Liu.
\newblock Octree-based fusion for realtime 3d reconstruction.
\newblock {\em Graph. Model.}, 75(3):126--136, 2013.

\bibitem{zollhofer2018state}
Michael Zollh{\"o}fer, Patrick Stotko, Andreas G{\"o}rlitz, Christian Theobalt,
  Matthias Nie{\ss}ner, Reinhard Klein, and Andreas Kolb.
\newblock State of the art on 3d reconstruction with rgb-d cameras.
\newblock In {\em Computer graphics forum}, pages 625--652. Wiley Online
  Library, 2018.

\end{thebibliography}
}

\end{document}